\documentclass[runningheads]{llncs}
\usepackage[T1]{fontenc}
\usepackage{graphicx}
\usepackage{subfig}
\usepackage{comment}
\usepackage{tcolorbox}
\usepackage{hyperref}  

\begin{document}
\title{Advancing Hate Speech Detection with Transformers: Insights from the MetaHate}
%
%
\author{%
Santosh Chapagain\inst{1} \and
Shah Muhammad Hamdi\inst{1} \and
Soukaina Filali Boubrahimi\inst{1}
}

\authorrunning{S. Chapagain et al.}

\titlerunning{ }

\institute{%
Department of Computer Science, Utah State University,\\
4205 Old Main Hill, Logan, Utah 84322, USA\\
\email{\{santosh.chapagain, s.hamdi, soukaina.boubrahimi\}@usu.edu}
}

\maketitle              

\begin{center}
    {\color{red} \textit{This paper includes examples of harmful and offensive language used for research and illustrative purposes only.}}
\end{center}

\begin{abstract}
Hate speech is a widespread and harmful form of online discourse, encompassing slurs and defamatory posts that can have serious social, psychological, and sometimes physical impacts on targeted individuals and communities. As social media platforms such as X (formerly Twitter), Facebook, Instagram, Reddit, and others continue to facilitate widespread communication, they also become breeding grounds for hate speech, which has increasingly been linked to real-world hate crimes. Addressing this issue requires the development of robust automated methods to detect hate speech in diverse social media environments. Deep learning approaches, such as vanilla recurrent neural networks (RNNs), long short-term memory (LSTM), and convolutional neural networks (CNNs), have achieved good results, but are often limited by issues such as long-term dependencies and inefficient parallelization. This study represents the comprehensive exploration of transformer-based models for hate speech detection using the MetaHate dataset—a meta-collection of 36 datasets with 1.2 million social media samples. We evaluate multiple state-of-the-art transformer models, including BERT, RoBERTa, GPT-2, and ELECTRA, with fine-tuned ELECTRA achieving the highest performance (F1 score: 0.8980). We also analyze classification errors, revealing challenges with sarcasm, coded language, and label noise.

\end{abstract}

\section{Introduction}
Hate speech is a particularly troubling type of offensive language that has been linked to real-world hate crimes on social media platforms \cite{muller2021fanning}. It can be a damaging form of online abuse, contributing to emotional distress, fostering social discrimination, and potentially inciting violence. Despite variations in definitions, there is consensus that hate speech targets individuals or groups with harmful intent, often based on race, religion, skin color, sexual identity, gender identity, ethnicity, disability, or national origin \cite{ala_hate_speech}.
The rise of social networks and online platforms has increased global connectivity but also highlighted a troubling surge in hate speech across geographical and cultural boundaries. In recent studies, approximately 30\% of the adolescents surveyed reported experiencing cyberbullying at some point in their lives. Furthermore, around 13\% indicated that they had been cyberbullied within the 30 days before the survey \cite{hinduja_patchin_2023}. Despite ongoing efforts by social media platforms to improve policies and detection systems, effectively addressing hate speech remains challenging due to complex and varying contexts, coded language, indirect expressions, and evolving slang \cite{elsherief2018hate}. The need for effective hate speech detection is critical. Advances in artificial intelligence (AI) and machine learning (ML) have shown promise in this area \cite{elsherief2018hate}, emphasizing the importance of capturing the context of hate speech. LLMs like BERT \cite{devlin2018bert}, based on the transformer architecture \cite{vaswani2017attention}, represent significant progress in NLP by utilizing attention mechanisms to capture longer dependencies. These models have excelled in tasks such as text classification and sentiment analysis \cite{fields2024survey}. However, they require large datasets. In this paper, we conduct a large-scale investigation into the contextual detection of hate speech using LLMs, using the extensive MetaHate \cite{piot2024metahate} dataset, which explored over 60 hate speech detection datasets and integrated 36 of them into the meta-collection. Our study presents a comprehensive evaluation of the transformer-based large language model for online hate detection. We evaluated several state-of-the-art models, including BART, ELECTRA, BERT, RoBERTa, and GPT-2 using the MetaHate dataset. Among the evaluated models, ELECTRA achieved the highest F1 score, outperforming all other baselines in hate speech classification.

\section{Related Work}
\subsection{Social Media as a Key Resource for Hate Speech Data}
Social media platforms like Reddit generate vast user content and have become key venues for hate speech \cite{mondal2017measurement}. Their anonymity enables the spread of hate, often linked to real-world crimes \cite{muller2021fanning}, and hate groups use coded symbols (e.g., "88" for "Heil Hitler") to organize \cite{league2019online}. Social media platforms facilitate the spread of online hate by making it easier to connect with people who share similar opinions, ideas, or interests, and by identifying potential targets based on searchable traits such as religion, gender, and race \cite{walther2022social}. For example, Twitter allows users to easily identify people's religion, gender, race, and political views through hashtags, profiles, and images. Social networks provide feedback mechanisms such as comments, likes, and upvotes that reinforce hateful behavior \cite{walther2022social}. These interactions can increase the frequency and intensity of hate messages, as well as strengthen the beliefs of those who spread them \cite{walther2022social}. Although platforms like Twitter and Facebook use human annotators to remove hateful content and encourage users to report it, these methods are labor-intensive and prone to bias. This has led to the development of automated hate speech detection using classical and deep learning algorithms. However, there is still an urgent need to develop robust models for real-time hate speech detection. Thus, the insights gained from the hate speech data on social networks are invaluable to developing more effective detection algorithms.

\subsection{Machine Learning Approaches in Hate Speech Detection}
Research in hate speech detection on social media has evolved from traditional machine learning methods like Support Vector Machines (SVM) and Logistic Regression to recent deep learning approaches such as LSTMs, CNNs, and more recently, transformer-based models like BERT \cite{devlin2018bert}. Warner and Hirschberg were among the first to apply supervised machine learning techniques to hate speech detection using SVM \cite{warner2012detecting}. Later, Davidson et al. \cite{davidson2017automated} expanded the use of algorithms in this domain with Logistic Regression, Naïve Bayes, Decision Trees, Random Forests, and linear SVMs. Naïve Bayes and K-NN classifiers have also proven effective, as demonstrated by Yan et al. \cite{yan2021stochastic} and Yin et al. \cite{yin2021slangs}, who combined these classifiers with techniques like bag-of-words and term frequency-inverse document frequency (TF-IDF). Kumari et al. \cite{kumari2021multi} proposed an LSTM-based approach for detecting hate speech on Twitter, demonstrating superior performance compared to shallow learning methods. Although deep learning and word embeddings-based models have shown promising performance, they could not match the capabilities of modern transformer-based models like BERT \cite{devlin2018bert} and RoBERTa \cite{liu2019roberta}. Mittal et al. \cite{mittal2023detecting} demonstrated that BERT-based models outperform traditional techniques by effectively capturing complex semantics and adapting to various linguistic contexts. LLMs has achieved remarkable performance in hate speech detection and other NLP tasks \cite{fields2024survey,cascalheira2023predicting,cascalheira2024lgbtq,chapagain2024predictive}  In this paper, we explore various transformer-based large language models, comparing the performance of fine-tuned transformers on the MetaHate \cite{piot2024metahate} dataset. Our study finds that ELECTRA outperforms other transformers on the MetaHate \cite{piot2024metahate} dataset.

\section{Big Data and Label}
The creators of the MetaHate \cite{piot2024metahate} dataset conducted a thorough review of over 60 existing hate speech detection datasets and ultimately selected 36 of them to include in their meta-collection. This process resulted in 1,667,496 entries, with 1,226,202 being unique comments sourced from social media platforms like Twitter, Facebook, and Reddit. The dataset excludes synthetic data and comments from non-social media sources. The dataset is binary-classified into hate (label=1) and non-hate (label=0) categories to align with the majority of existing datasets and facilitate ease of use. Of the 1,226,202 comments, 253,145 are labeled as hate speech, representing 20.64\% of the dataset. To address the class imbalance in the MetaHate \cite{piot2024metahate} dataset, we used class weights from scikit-learn. We observed that hate speech contains a high frequency of terms such as "fucking", "faggot", "nigger", "black", "idiot", and others. 

\begin{figure}[!htb]
    \centering
    \subfloat[\centering Hate (label = 1)]{{\includegraphics[width=4.0cm]{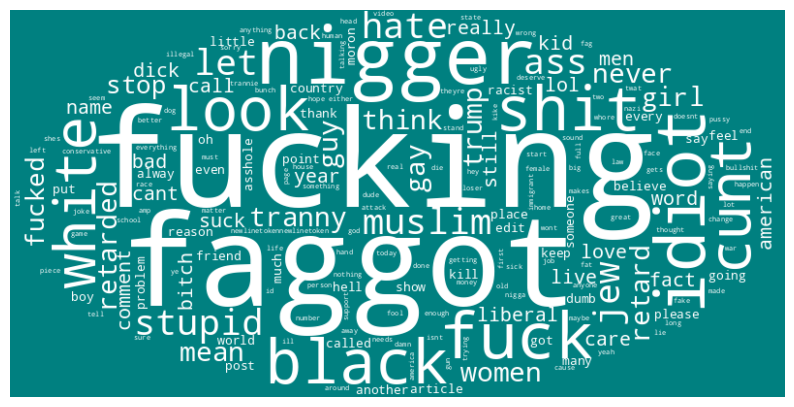} }}%
    \qquad
    \subfloat[\centering No Hate (label = 0 )]{{\includegraphics[width=4.0cm]{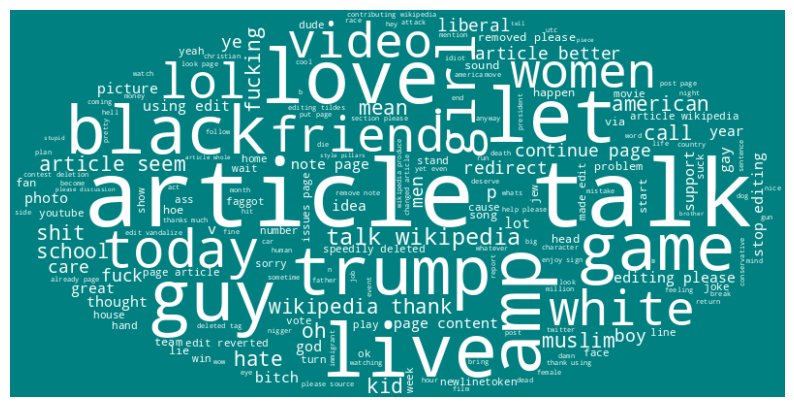} }}%
    \caption{WordCloud on MetaHate \cite{piot2024metahate} dataset labels.}%
    \label{fig:example2}%
\end{figure}

\section{ELECTRA}
In this study, we used ELECTRA (Efficiently Learning an Encoder that Classifies Token Replacements Accurately), a transformer-based architecture that pretrains two models: a generator and a discriminator. The generator acts as a masked language model by replacing tokens in a sequence, while the discriminator identifies these replaced tokens. In the fine-tuning step, the preprocessed data were split as follows: 80\% of instances were randomly allocated for fine-tuning and training, 10\% for validation, and 10\% for testing the final model's performance. Fine-tuning began with tokenization, where the input text is broken down into individual tokens. These tokens were then passed sequentially as input embeddings, undergoing multi-head self-attention after adding positional embeddings. In the self-attention mechanism, the model computed query, key, and value vectors for each token to capture dependencies across the sequence, allowing it to understand relationships between tokens regardless of their position. These representations were then processed by a feedforward network to introduce non-linear transformations. After passing through stacked layers of encoder blocks, the output sequence of vectors from ELECTRA was passed through a linear layer, dropout layer, and softmax layer to distinguish between classes.

\section{Experiments}
We ran all experiments on a Linux server with dual Intel Xeon Gold 5220R processors (24 cores each, 2.20 GHz) and eight NVIDIA RTX A4000 GPUs (16 GB VRAM each). Training was carried out using PyTorch 2.4.0 with CUDA 12.1 for all models. All of our code is available on GitHub \footnote{\url{https://github.com/chapagaisa/hate_speech_detection}}. We fine-tuned the models for 4 epochs with a batch size of 32 and a maximum sequence length of 300, chosen empirically for efficiency without loss in performance. Training used the AdamW optimizer (learning rate 3e-5, epsilon 1e-6, weight decay 0.001), excluding bias and LayerNorm from decay.

\subsection{Baseline Methods}
All baseline methods used consistent hyperparameters (see Section V) and relied on base versions of transformer models for fair comparison. Traditional models included a linear SVM with TF-IDF and stop word removal, and a CNN trained for one epoch on padded tokenized sentences using a simple architecture with ReLU and sigmoid layers. Transformer-based models included BERT, RoBERTa, GPT-2, BART, DeBERTa, Longformer, XLNet, DistilBERT, and T5. These models vary in architecture and training strategies: RoBERTa improves on BERT by removing next-sentence prediction, GPT-2 is a unidirectional transformer, BART combines bidirectional and autoregressive features, DeBERTa introduces disentangled attention, Longformer handles long sequences efficiently, XLNet uses an autoregressive pretraining method, DistilBERT is a compact version of BERT, and T5 reformulates all tasks into a text-to-text format.

\subsection{Results and Analysis}
Table 1 presents the results of various transformer-based models fine-tuned with the MetaHate dataset. ELECTRA performed the best among the baselines, achieving the highest F1 score of 0.8980 and accuracy of 0.8946, while the worst-performing model was GPT-2, with an F1 score of 0.6504 and accuracy of 0.6152. DistilBERT and BART showed competitive results based on F1 score and accuracy. In the mid-tier, RoBERTa and XLNet showed similar results, slightly below the top models.

\begin{table}[!htbp]
\centering
\caption{Performance of classifiers on MetaHate \cite{piot2024metahate}}
\begin{tabular}{p{3cm} p{2cm} p{2cm}}
\hline 
\textbf{Model} & \textbf{F1 Score} & \textbf{Accuracy} \\
\hline 
SVM & 0.8380 & 0.8466 \\
CNN & 0.8422 & 0.8612 \\
BERT & 0.8809 & 0.8879 \\
GPT2 & 0.6504 & 0.6152 \\
T5 & 0.8707 & 0.8625 \\
DeBERTa & 0.8808 & 0.8746 \\
Longformer & 0.8845 & 0.8785 \\
RoBERTa & 0.8908 & 0.8858 \\
XLNet & 0.8917 & 0.8870 \\
BART & 0.8928 & 0.8886 \\
DistilBERT & 0.8940 & 0.8905 \\
ELECTRA & \textbf{0.8980} & \textbf{0.8946} \\
\hline
\end{tabular}
\end{table}

Figure 2 shows the confusion matrix for the ELECTRA model on the MetaHate test set. Out of 110,117 instances, the model correctly classified 78,627 non-hate samples (71.4\%) and 19,882 hate speech samples (18.1\%). Misclassifications included 8,215 false positives (7.5\%) and 3,393 false negatives (3.1\%). These results indicate a strong overall performance with high precision and recall, especially in correctly identifying hate speech. Figure 3 shows the ROC curves and AUC scores of various transformer-based classifiers. ELECTRA's ROC-AUC is 0.9533, slightly below RoBERTa and XLNet. Although ROC-AUC is an important metric for illustrating classifier accuracy by distinguishing between positive and negative classes across diverse thresholds, the F1 score is more attuned to class imbalance. ELECTRA outperforms all classifiers in terms of F1 score, indicating their excellence in correctly identifying hate speech instances and effectively addressing class imbalance.

\begin{figure*}[htbp]
\centering
\begin{minipage}{0.48\textwidth}
    \centering
    \includegraphics[width=\linewidth]{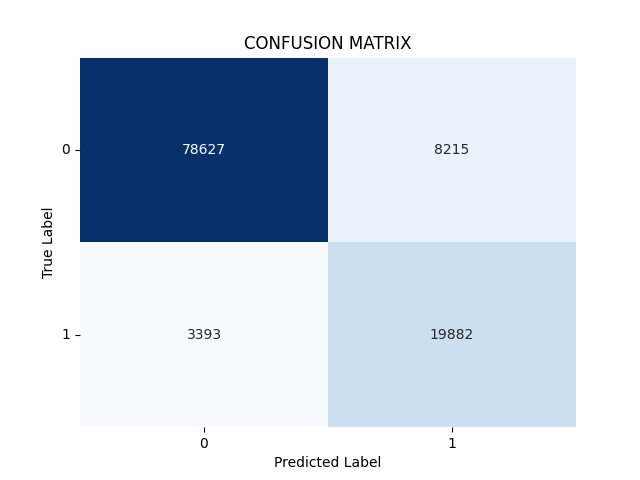}
    \caption{Confusion matrix on the test set using ELECTRA.}
    \label{fig:electra_confusion}
\end{minipage}
\hfill
\begin{minipage}{0.48\textwidth}
    \centering
    \includegraphics[width=\linewidth]{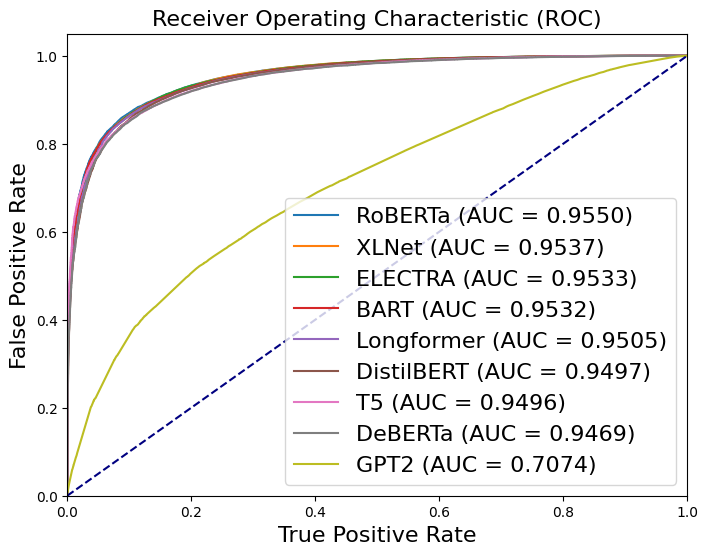}
    \caption{ROC-AUC of different transformers on MetaHate dataset \cite{piot2024metahate}.}
    \label{fig:roc_auc}
\end{minipage}
\end{figure*}

\subsubsection{Linguistic Analysis of Misclassifications}
We performed a linguistic analysis on some misclassified examples of the ELECTRA classifier and found several recurring patterns that included sarcastic or ironic language (e.g., remarks that mimic exaggerated praise or mock stereotypes) were flagged as hateful despite lacking genuine hostility. Another example is figurative and exaggerated expressions, such as dramatic warnings or over-the-top threats delivered in jest, which were also misinterpreted due to their intense tone. Similarly, keyword bias, labeling emotionally charged but harmless expressions (e.g., frustration over a delayed service) caused the model to classify hate as non-hateful based on strong wording. In contrast, aggressive messages without direct hate indicators such as all-caps complaints or accusatory statements were sometimes missed. 

\section{DISCUSSION AND CONCLUSION}
Although ELECTRA achieved the highest F1 score among the tested models, a small but significant proportion of MetaHate samples may be mislabeled, particularly around subjective or political discourse. We advocate for semi-automated relabeling tools using explainable AI (e.g., LLM-based rationale classification) to improve dataset integrity.  Future work should explore hybrid approaches combining fine-tuning with instruction or adapter-based methods for low-resource and cross-platform settings. Developing efficient transformer variants can support real-time use on edge devices. Expanding to multi-modal hate content and improving annotation quality for evaluations are also key directions.

Addressing online hate speech involves ethical considerations, particularly around free speech. In our work, we used the publicly available MetaHate dataset, which was curated from multiple sources without including any personally identifiable information. The dataset supports the detection of harmful content on social media but contains offensive elements, so careful handling is necessary to prevent misuse. Some parts of the dataset are openly accessible, while others require author permission. Any biases present are from the original sources and are not intended to cause harm. We used the dataset for only research purposes, which is in line with its intended non-commercial use.

This study demonstrates the effectiveness of transformer-based models, particularly ELECTRA, in detecting hate speech within a large-scale and diverse dataset. We also performed detailed analyses of misclassified instances and found common linguistic challenges, such as sarcasm, coded language, and annotation noise, that continue to hinder classification accuracy. By integrating insights from NLP, social science, and ethics, we advance toward more robust and socially responsible AI systems to combat online hate.

\section{Acknowledgements}
Shah Muhammad Hamdi and Soukaina Filali Boubrahimi are supported by the CISE and GEO directorates under NSF awards \#2204363, \#2240022, \#2301397, and \#2305781.

%
%
%
%
\bibliographystyle{splncs04}
\bibliography{samplepaper}

\end{document}